\documentclass[preprint,12pt]{elsarticle}
\makeatletter
\def\ps@pprintTitle{%
 \let\@oddhead\@empty
 \let\@evenhead\@empty
 \def\@oddfoot{}%
 \let\@evenfoot\@oddfoot}
\makeatother

\usepackage{amssymb}
\usepackage{verbatim}
\usepackage{float}
\usepackage{epstopdf}
\usepackage{algorithm,algcompatible,amsmath}
\usepackage{gensymb}
\usepackage{booktabs}
\usepackage{amssymb,verbatim,epstopdf,algorithm,algcompatible,amsmath,caption,subcaption,algcompatible}
\usepackage{xcolor}
\usepackage{subcaption}
\usepackage{algcompatible}
\usepackage[flushleft]{threeparttable}

\newcommand\mynotes[1]{\textcolor{red}{#1}}
\newcommand\tgknotes[1]{\textcolor{green}{#1}}
\renewcommand\mynotes[1]{} 
\renewcommand\tgknotes[1]{} 

\begin{document}
\begin{frontmatter}
\title{Experience Replay Using Transition Sequences}

\author{Thommen George Karimpanal, Roland Bouffanais}
\ead{thommen\_george@mymail.sutd.edu.sg, bouffanais@sutd.edu.sg}
\address{Singapore University of Technology and Design, 8 Somapah Road, Singapore 487372}

\begin{abstract}

    Experience replay is one of the most commonly used approaches to improve the sample efficiency of reinforcement learning algorithms. In this work, we propose an approach to
    select and replay sequences of transitions in order to accelerate the
    learning of a reinforcement learning agent in an off-policy setting.  In
    addition to selecting appropriate sequences, we also artificially
    construct transition sequences using information gathered from previous
    agent-environment interactions. These sequences, when replayed, allow
    value function information to trickle down to larger sections of the
    state/state-action space, thereby making the most of the agent's
    experience.
    We demonstrate our approach on modified versions of standard reinforcement
    learning tasks such as the mountain car and puddle world problems and
    empirically show that it enables faster, and more accurate learning of value functions as
    compared to other forms of experience replay.
    Further, we briefly discuss some of the possible extensions to this work,
    as well as applications and situations where this approach could be
    particularly useful.  
\end{abstract}

\end{frontmatter}

\section{Introduction}

\label{intro}
Real-world artificial agents ideally need to be able to learn as much as
possible from their interactions with the environment. This is especially true
for mobile robots operating within the reinforcement learning (RL) framework,
where the cost of acquiring information from the environment through
exploration generally exceeds the computational cost of
learning ~\citep{wang2016sample,adam2012experience,schaul2015prioritized}.

Experience replay \citep{lin1992self} is a technique that reuses information
gathered from past experiences to improve the efficiency of learning.
In order to replay stored experiences using this approach, an off-policy
\citep{sutton2011reinforcement,geist2014off} setting is a prerequisite. In
off-policy learning, the policy that dictates the agent's control actions is
referred to as the behavior policy. Other policies corresponding to the value/action-value functions of different tasks that the agent aims to learn are referred to as target policies.
Off-policy algorithms
utilize the agent's behavior policy to interact with the environment, while
simultaneously updating the value functions associated with the target
policies. These algorithms can hence be used to parallelize learning, and,
thus gather as much knowledge as possible using real experiences
\citep{sutton2011horde,white2012scaling,modayil2014multi}. However, when the
behavior and target policies differ considerably from each other, the actions
executed by the behavior policy may only seldom correspond to those recommended by the
target policy. This could lead to poor estimates of the corresponding value
function.  Such cases could arise in multi-task scenarios where multiple tasks are learned in an off-policy manner. Also, in general, in environments where desirable
experiences are rare occurrences, experience replay could be employed to
improve the estimates by storing and replaying transitions (state, actions and
rewards) from time to time.

Although most experience replay approaches store and reuse individual transitions,
replaying sequences of transitions could offer certain advantages. For instance, if a value function
update following a particular transition results in a relatively large change
in the value of the corresponding state or state-action pair, this change will
have a considerable influence on the bootstrapping targets of states or
state-action pairs that led to this transition. Hence, the effects of this
change should ideally be propagated to these states or state-action pairs. If
instead of individual transitions, sequences of transitions are replayed, this
propagation can be achieved in a straightforward manner.  Our approach aims to
improve the efficiency of learning by replaying transition sequences in this
manner. The sequences are selected on the basis of the magnitudes of the
temporal difference (TD) errors\citep{sutton2011reinforcement}, associated with them. We hypothesize that
selecting sequences that contain transitions associated with higher magnitudes
of TD errors allow considerable learning progress to take place. This is
enabled by the propagation of the effects of these errors to the values
associated with other states or state-action pairs in the transition sequence.

Replaying a larger variety of such sequences would result in a more efficient
propagation of the mentioned effects to other regions in the
state/state-action space. Hence, in order to aid the propagation in this
manner, other sequences that could have occurred are artificially constructed
by comparing the state trajectories of previously observed sequences. These
virtual transition sequences are appended to the replay memory, and they help
bring about learning progress in other regions of the state/state-action space
when replayed.
\begin{figure}[ht]
  \centering
  \includegraphics[width=1\linewidth]{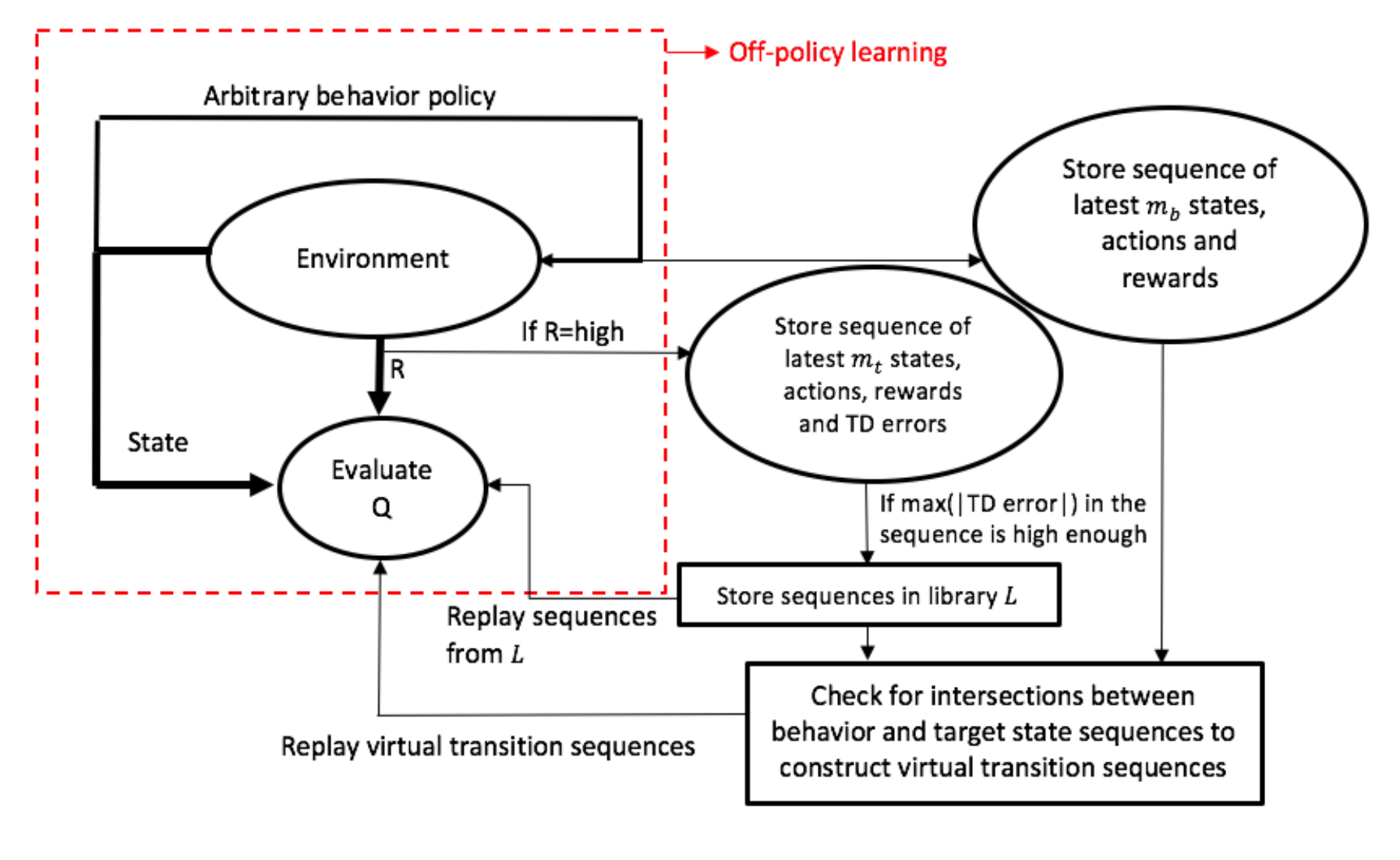}
  \caption{Structure of the proposed algorithm in contrast to the traditional
    off-policy structure. $Q$ and $R$ denote the action-value function
    and reward respectively.}
  \label{fig:offpolmodified}
\end{figure}
The generated transition sequences are virtual in the sense that they may have
never occurred in reality, but are constructed from sequences that have
actually occurred in the past. 
The additional replay updates corresponding to the mentioned transition sequences supplement the regular off-policy value function updates that follow the real-world execution of actions, thereby making the most out of the agent's interactions with the environment.

\section{Related Work}
\label{background}

The problem of learning from limited experience is not new in the field of RL
\citep{thrun1992efficient, thomas2016data}. Generally, learning speed and
sample efficiency are critical factors that determine the feasibility of
deploying learning algorithms in the real
world. 
Particularly for robotics applications, these factors are even more important,
as exploration of the environment is typically time and energy expensive
\citep{bakker2006quasi,kober2013reinforcement}. It is thus important for a
learning agent to be able to gather as much relevant knowledge as possible
from whatever exploratory actions occur.

Off-policy algorithms are well suited to this need as it enables multiple
value functions to be learned together in parallel.
When the behavior and target policies vary considerably from each other,
importance sampling \citep{rubinstein2016simulation,sutton2011reinforcement} is
commonly used in order to obtain more accurate estimates of the value
functions. Importance sampling reduces the variance of the estimate by taking
into account the distributions associated with the behavior and target
policies, and making modifications to the off-policy update equations
accordingly.  However, the estimates are still unlikely to be close to their
optimal values if the
agent receives very little experience relevant to a particular task.

This issue is partially addressed with experience replay, in which information contained in the replay memory is used from time to time in order to update the value functions. As a result, the agent is able to learn from uncorrelated historical data, and the sample efficiency of learning is greatly improved.
This approach has received a lot of attention in recent years due to its
utility in deep RL applications
\citep{mnih2015human,mnih2016asynchronous,mnih2013playing,adam2012experience,de2015importance}.
Recent works~\citep{schaul2015prioritized,NarasimhanKB15} have revealed that 
certain transitions are more useful
than others. 
Schaul
\textit{et al.}~\citep{schaul2015prioritized} prioritized transitions on the
basis of their associated TD errors. They also briefly mentioned the
possibility of replaying transitions in a sequential manner. The experience
replay framework developed by Adam \textit{et al.}~\citep{adam2012experience}
involved some variants that replayed sequences of experiences, but these
sequences were drawn randomly from the replay memory. More recently, Isele et al. \citep{isele2018selective} reported a selective experience replay approach aimed at performing well in the context of lifelong learning \citep{thrun1996learning}. The authors of this work proposed a long term replay memory in addition to the conventionally used one. Certain bases for designing this long-term replay memory, such as favoring transitions associated with high rewards and high absolute TD errors are similar to the ones described in the present work. However, the approach does not explore the replay of sequences, and its fundamental purpose is to shield against catastrophic forgetting \citep{goodfellow2013empirical} when multiple tasks are learned in sequence. The replay approach described in the present work focuses on enabling more sample-efficient learning in situations where positive rewards occur rarely. Apart from this, Andrychowicz et al. \citep{andrychowicz2017hindsight}  proposed a hindsight experience replay approach, directed at addressing this problem, where each episode is replayed with a goal that is different from the original goal of the agent. The authors reported significant improvements in the learning performance in problems with sparse and binary rewards. These improvements were essentially brought about by allowing the learned value/$Q$ values (which would otherwise remain mostly unchanged due to the sparsity of rewards) to undergo significant change under the influence of an arbitrary goal. The underlying idea behind our approach also involves modification of the $Q-$values in reward-sparse regions of the state-action space. The modifications, however, are not based on arbitrary goals, and are selectively performed on state-action pairs associated with successful transition sequences associated with high absolute TD errors. Nevertheless, the hindsight replay approach is orthogonal to our proposed approach, and hence, could be used in conjunction with it.

Much like in Schaul \textit{et al.} \citep{schaul2015prioritized}, TD errors have been frequently used as a basis for prioritization in other RL problems
\citep{white2014surprise,Thrun92efficientexploration,schaul2015prioritized}. In
particular, the model-based approach of prioritized sweeping
\citep{moore1993prioritized,ICML-c3-2013-SeijenS} prioritizes backups that are
expected to result in a significant change in the value function.
The algorithm we propose here uses a model-free architecture, and it is based on the idea of
selectively reusing previous experience. However, we describe the reuse of
sequences of transitions based on the TD errors observed when these
transitions take place.  Replaying sequences of experiences also seems to be
biologically plausible
\citep{olafsdottir2015hippocampal,buhry2011reactivation}. In addition, it is
known that animals tend to remember experiences that lead to high rewards
\citep{singer2009rewarded}. This is an idea reflected in our work, as only
those transition sequences that lead to high rewards are considered for being
stored in the replay memory. In filtering transition sequences in this manner,
we simultaneously address the issue of determining which experiences are to be
stored.

In addition to selecting transition sequences, we also generate virtual sequences of transitions which the agent could have possibly experienced, but in reality, did not. This virtual experience is then replayed to improve the
agent's learning.  Some early approaches in RL, such as the dyna architecture
\citep{sutton1990integrated} also made use of simulated experience to improve
the value function estimates. However, unlike the approach proposed here, the
simulated experience was generated based on models of the reward function and
transition probabilities which were continuously updated based on the agent's
interactions with the environment. In this sense, the virtual experience
generated in our approach is more grounded in reality, as it is based directly
on the data collected through the agent-environment interaction. In more recent work, Fonteneau \textit{et al.} describe an approach to generate artificial trajectories and use them to find policies with acceptable performance guarantees \citep{fonteneau2013batch}. However, this approach is designed for batch RL, and the generated artificial trajectories are not constructed using a TD error basis. Our approach
also recognizes the real-world limitations of replay memory
\citep{de2015importance}, and stores only a certain amount of information at a
time, specified by memory parameters. The selected and generated sequences are
stored in the replay memory in the form of libraries which are continuously
updated so that the agent is equipped with transition sequences that are most
relevant to the task at hand.





\section{Methodology}
\label{method}


The idea of selecting appropriate transition sequences for replay is
relatively straightforward. In order to improve the agent's learning, first,
we simply keep track of the state, actions, rewards and absolute values of the TD errors associated
with each transition. Generally, in difficult learning environments, high rewards occur rarely. So, when such an
event is observed, we consider storing the corresponding sequence of
transitions into a replay library $L$. In this manner, we use the reward
information as a means to filter transition sequences. The approach is similar
to that used by Narasimhan \textit{et al.}~\citep{NarasimhanKB15}, where
transitions associated with positive rewards are prioritized for replay.

Among the transition sequences considered for inclusion in the library $L$,
those containing transitions with high absolute TD error values are considered
to be the ones with high potential for learning progress. Hence, they are
accordingly prioritized for replay. The key idea is that when the TD error
associated with a particular transition is large in magnitude, it generally
implies a proportionately greater change in the value of the corresponding
state/state-action pair. Such large changes have the potential to influence
the values of the states/state-action pairs leading to it, which implies a
high potential for learning. Hence, prioritizing such sequences of transitions
for replay is likely to bring about greater learning progress.  
Transition sequences associated with large magnitudes of TD error are retained
in the library, while those with lower magnitudes are removed and replaced
with superior alternatives. In reality, such transition sequences may be very
long and hence, impractical to store. Due to such practical considerations, we
store only a portion of the sequence, based on a predetermined memory
parameter.  The library is continuously updated as and when the
agent-environment interaction takes place, such that it will eventually
contain sequences associated with the highest absolute TD errors.

As described earlier, replaying suitable sequences allows the effects of large
changes in value functions to be propagated throughout the sequence. In order
to propagate this information even further to other regions of the
state/state-action space, we use the sequences in $L$ to construct additional
transition sequences which could have possibly occurred. These virtual
sequences are stored in another library $L_{v}$, and later used for experience
replay.

\begin{figure}[h!]
\begin{center}
\includegraphics[width=15cm]{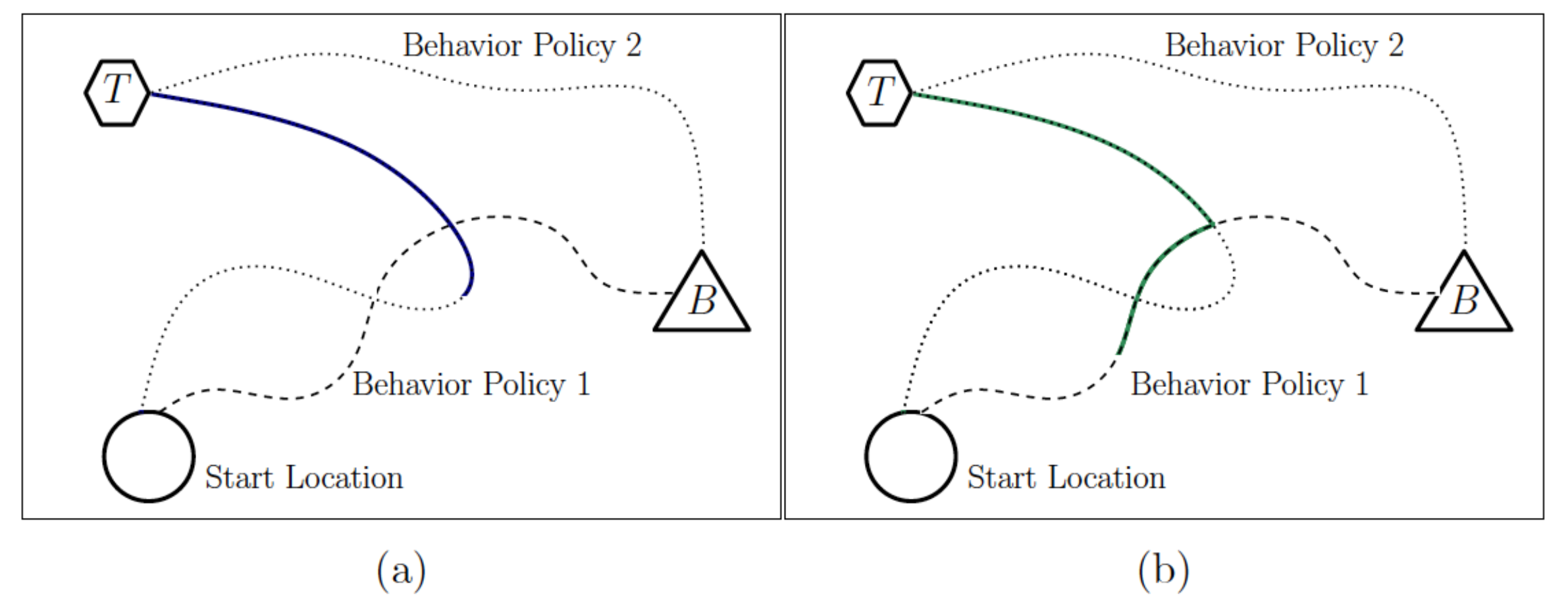}
\end{center}
\caption{(a) Trajectories corresponding to two hypothetical behavior policies are shown. A portion of the trajectory associated with a high reward (and stored in $L$) is highlighted (b) The virtual trajectory constructed from the two behavior policies is highlighted. The states, actions and rewards associated with this trajectory constitute a virtual transition sequence.}
\label{fig:overlap}
\end{figure}

In order to intuitively describe our approach of artificially constructing
sequences, we consider the hypothetical example shown in Figure
\ref{fig:overlap}(a), where an agent executes behavior policies that help it
learn to navigate towards location $B$ from the start location. However, using
off-policy learning, we aim to learn value functions corresponding to the
policy that helps the agent navigate towards location $T$.

The trajectories shown in Figure \ref{fig:overlap}(a) correspond to hypothetical
actions dictated by the behavior policy midway through the learning process,
during two separate episodes. The trajectories begin at the start location and
terminate at location $B$. However, the trajectory corresponding to behavior
policy $2$ also happens to pass through location $T$, at which point the agent
receives a high reward. This triggers the transition sequence storage
mechanism described earlier, and we assume that some portion of the sequence
(shown by the highlighted portion of the trajectory in Figure
\ref{fig:overlap}(a)) is stored in library $L$.  Behavior policy $1$ takes the
agent directly from the start location towards the location $B$, where it
terminates. As the agent moves along its trajectory, it intersects with the
state trajectory corresponding to the sequence stored in
$L$. 
Using this intersection, it is possible to artificially construct additional
trajectories (and their associated transition sequences) that are successful
with respect to the task of navigating to location $T$. The highlighted
portions of the trajectories corresponding to the two behavior policies in
Figure \ref{fig:overlap}(b) show such a state trajectory, constructed using
information related to the intersection of portions of the two previously
observed
trajectories. 
The state, action and reward sequences associated with this highlighted
trajectory form a virtual transition sequence.

Such artificially constructed transition sequences present the possibility of
considerable learning progress. This is because, when replayed, they help
propagate the large learning potential (characterized by large magnitudes of
TD errors) associated with sequences in $L$ to other regions of the
state/state-action space.  These replay updates supplement the off-policy
value function updates that are carried out in parallel, thus accelerating the
learning of the task in question. This outlines the basic idea behind our
approach.


Fundamentally, our approach can be decomposed into three steps:
\begin{enumerate}
\item Tracking and storage of relevant transition sequences
\item Construction of virtual transition sequences using the stored transition
  sequences
\item Replaying the transition sequences
\end{enumerate}
These steps are explained in detail in Sections \ref{subsect1}, \ref{subsect2}
and \ref{subsect3}.

\subsection{Tracking and Storage of Relevant Transition Sequences}
\label{subsect1}

As described, virtual transition sequences are constructed by joining together
two transition sequences. One of them, say $\Theta_{t}$, composed of $m_{t}$
transitions, is historically successful---it has experienced high rewards with
respect to the task, and is part of the library $L$. The other sequence,
$\Theta_{b}$, is simply a sequence of the latest $m_{b}$ transitions executed
by the agent.

If the agent starts at state $s_{0}$ and moves through intermediate states
$s_{i}$ and eventually to $s_{j+1}$ (most recent state) by executing a series of
actions $a_{0}...a_{i}...a_{j}$, it receives rewards
${R}_{0}...{R}_{i}...{R}_{j}$ from the environment. These
transitions comprise the transition sequence $\Theta_{b}$.
\\ \\
\begin{equation}\label{behaviortransitionseq_eqn}
  \Theta_{b}=\begin{cases}
    [S(0:j)\quad \pi(0:j)\quad{R}(0:j)]\qquad \qquad \qquad \text{if}\quad j\leq m_{b} \\ [S((j-m_{b}):j)\quad \pi((j-m_{b}):j)\quad{R}((j-m_{b}):j)]\quad \text{otherwise}\end{cases}
\end{equation}
where:
\begin{align*}
S(x:y)&=(s_{x}...s_{i}...s_{y}),\\
\pi(x:y)&=(a_{x}...a_{i}...a_{y}),\\
{R}(x:y)&=({R}_{x}...{R}_{i}...{R}_{y}).
\end{align*}
We respectively refer to $S(x:y)$, $\pi(x:y)$ and ${R}(x:y)$ as
the state, action and reward transition sequences corresponding to a series of agent-environment interactions, indexed from $x$ to $y$ ($x,y\in \mathbb{N}$).

For the case of the
transition sequence $\Theta_{t}$, we keep track of the sequence of TD errors
$\delta_{0}...\delta_{i}...\delta_{k}$ observed as well. If a high reward is observed in transition $k$, then:
\begin{equation}\label{targettransitionseq_eqn}
  \Theta_{t}=\begin{cases}
    [S(0:k)\quad \pi(0:k)\quad{R}(0:k)\quad{\Delta(0:k)]}\qquad \qquad \qquad \text{if}\quad k\leq m_{t} \\ [S((k-m_{t}):k)\quad \pi((k-m_{t}):k)\quad{R}((k-m_{t}):k)\quad \Delta((k-m_{t}):k)]\quad \text{otherwise}\end{cases}
\end{equation}
where $\Delta(x:y)=(|\delta_{x}|...|\delta_{i}|...|\delta_{y}|)$.

The memory parameters $m_{b}$ and $m_{t}$ are chosen based on the memory
constraints of the agent. They determine how much of the recent
agent-environment interaction history is to be stored in memory.

It is possible that the agent encounters a number of transitions associated
with high rewards while executing the behavior policy. Corresponding to these transitions, a number
of successful transition sequences $\Theta_{t}$ would also exist.
These sequences are maintained in the library $L$ in a manner similar to the
\textit{Policy Library through Policy Reuse (PLPR)} algorithm~\citep{fernandez2005building}. To decide whether to include a
new transition sequence ${\Theta_{t}}_{\text{\scriptsize new}}$ into the library $L$, we
determine the maximum value of the absolute TD error sequence $\Delta$
corresponding to ${\Theta_{t}}_{\text{\scriptsize new}}$ and check whether it
is $\tau$-close---the parameter $\tau$ determines the exclusivity of the library---to the maximum of the corresponding values associated with the transition
sequences in $L$. If this is the case, then ${\Theta_{t}}_{\text{\scriptsize new}}$ is included
in $L$. Since the transition sequences are filtered based on the maximum of the absolute values of TD errors among all the transitions in a sequence, this approach should be able to mitigate problems stemming from low magnitudes of TD errors associated with local optima \citep{baird1999reinforcement,tutsoy2016chaotic}. Using the absolute TD error as a basis for selection, we maintain a fixed number ($l$) of
transition sequences in the library $L$. This ensures that the library is
continuously updated with the latest transition sequences associated with the
highest absolute TD errors. The complete algorithm is illustrated in Algorithm
\ref{alg:algorithm1}.

\begin{algorithm}[h!]
  \caption{Maintaining a replay library of transition sequences}
  \begin{algorithmic}[1]
    \STATE \textbf{Inputs}: \STATEx $\tau:$~Parameter that determines the
    exclusivity of the library \STATEx $l:$~Parameter that determines the
    number of transition sequences allowed in the library
    \STATEx $\Delta_{k}:$ Sequence of absolute TD errors corresponding to a transition sequence $\Theta_{k}$
    \STATEx
    $L=\{{\Theta_{t}}_{0}...{\Theta_{t}}_{i}...{\Theta_{t}}_{m}\}:$ A library
    of transition sequences ($m\leq l$) \STATEx ${\Theta_{t}}_{\text{\scriptsize new}}:$~New
    transition sequence to be evaluated
    \STATE $W_{\text{\scriptsize new}}=\max({\Delta_{t}}_{\text{\scriptsize new}})$
    \FOR {$j=1:m$}
    \STATE $W_{j}=\max({\Delta_{t}}_{j}$) 
    \ENDFOR
    \IF {$W_{\text{\scriptsize new}}*\tau>\max(W)$} \STATE $L=L\cup\{{\Theta_{t}}_{\text{\scriptsize new}}\}$ \STATE
    $n_{t}=$Number of transition sequences in $L$ \IF {$n_{t}>l$} \STATE
    $L=\{{\Theta_{t}}_{n_{t}-l}...{\Theta_{t}}_{i}...{\Theta_{t}}_{n_{t}}\}$
    \ENDIF
    \ENDIF
  \end{algorithmic}
  \label{alg:algorithm1}
\end{algorithm}

\subsection{Virtual Transition Sequences}
\label{subsect2}
Once the transition sequence $\Theta_{b}$ is available and a library $L$ of
successful transition sequences ${\Theta_{t}}$ is obtained, we use this
information to construct a library $L_{v}$ of virtual transition sequences
$\Theta_{v}$. The virtual transition sequences are constructed by first
finding points of intersection $s_{c}$ in the state transition sequences of
$\Theta_{b}$ and the $\Theta_{t}$'s in $L$.

Let us consider the transition sequence $\Theta_{b}$:
\begin{align*}
  \Theta_{b}&=[{S(x:y)\quad \pi(x:y)\quad{R}(x:y)]},
\intertext{and a transition sequence $\Theta_{t}$:}
  \Theta_{t}&=[{S(x':y')\quad    \pi(x':y')\quad{R}(x':y')\quad{\Delta(x':y')}]},
\intertext{Let $\Theta_{t_{s}}$ be a sub-matrix of $\Theta_{t}$ such that:}
\end{align*}
\begin{equation}
\label{eqn:thetaTs}
  \Theta_{t_{s}}=[{S(x':y')\quad \pi(x':y')\quad{R}(x':y')]},
\end{equation}
\begin{align*}
\intertext{Now, if $\sigma_{x}^{y}$ and $\sigma_{x'}^{y'}$ are sets containing all the elements of sequences $S(x:y)$ and $S(x':y')$ respectively, and if $\exists s_{c} \in \{\sigma_{x}^{y} \cap \sigma_{x'}^{y'} \}$, then:}
{S(x:y)}&=(s_{x},...s_{c},s_{c+1},...s_{y}),
\intertext{and}
{S(x':y')}&=(s_{x'}...s_{c},s_{c+1}...s_{y'}).
\end{align*}
Once points of intersection have been obtained as described above, each of the
two sequences $\Theta_{b}$ and $\Theta_{t_{s}}$ are decomposed into two subsequences at the point of intersection such that:
\begin{equation}
\label{eqn:theta_b_decom}
\Theta_{b}=
\begin{bmatrix}
    \Theta_{b}^{1} \\ \Theta_{b}^{2}
 \end{bmatrix}
\end{equation}

where

$\Theta_{b}^{1}=[{S(x:c)\quad \pi(x:c)\quad{R}(x:c)]}$
\newline
and

$\Theta_{b}^{2}=[{S((c+1):y)\quad
  \pi((c+1):y)\quad{R}((c+1):y)]}$  
\newline

Similarly,
\begin{equation}
\label{eqn:theta_ts_decom}
\Theta_{t_{s}}=
\begin{bmatrix}
    \Theta_{t_{s}}^{1} \\ \Theta_{t_{s}}^{2}
 \end{bmatrix}
\end{equation}
where 

$\Theta_{t_{s}}^{1}=[{S(x':c)\quad \pi(x':c)\quad{R}(x':c)]}$ 
\newline 
and

$\Theta_{t_{s}}^{2}=[{S((c+1):y')\quad
  \pi((c+1):y')\quad{R}((c+1):y')]}$ 
\newline

The virtual transition sequence is then simply:
\begin{equation}
\label{eqn:VTSfrombehandtargsequences}
\Theta_{v}=
\begin{bmatrix}
    \Theta_{b}^{1} \\ \Theta_{t_{s}}^{2}
 \end{bmatrix}
\end{equation}

We perform the above procedure for each transition sequence in $L$ to obtain
the corresponding virtual transition sequences ${\Theta_{v}}$. These virtual
transition sequences are stored in a library $L_{v}$:
\begin{align*}
  L_{v}=\{{\Theta_{v}}_{1}...{\Theta_{v}}_{i}...{\Theta_{v}}_{n_{v}}\},
\end{align*}
where ${n_{v}}$ denotes the number of virtual transition sequences in $L_{v}$,
subjected to the constraint ${n_{v}}\leq{l}$.

The overall process for constructing and storing virtual transition sequences is summarized in Algorithm \ref{alg:VTS}. Once the library $L_{v}$ has been constructed,
we replay the sequences contained in it to improve the estimates of the value function. The details of
this are discussed in Section \ref{subsect3}.

\begin{algorithm}[h!]
  \caption{Constructing virtual transition sequences}
  \begin{algorithmic}[1]
    \STATE \textbf{Inputs}: \STATEx Sequence of latest $m_{b}$ transitions $\Theta_{b}$ 
    \STATEx Library $L$ containing $n_{t}$ stored transition sequences 
     \STATEx Library $L_v$ for storing virtual transition sequences 
    \FOR {$t=1:n_{t}$}
    \STATE Extract $\Theta_{t_{s}}$ from $\Theta_{t}$ (Equation \ref{eqn:thetaTs})
    \STATE Find set of states $S_{I}$ corresponding to the intersection of the state trajectories of $\Theta_{b}$ and $\Theta_{t_{s}}$ 
    \IF {$S_{I}$ is not empty,}
    \FOR {each state $s_{i}$ in $S_{I}$,}
      \STATE Treat $s_{i}$ as the intersection point and decompose $\Theta_{b}$ and $\Theta_{t_{s}}$ as per Equations \ref{eqn:theta_b_decom} and \ref{eqn:theta_ts_decom}
      \ENDFOR
      \STATE Choose $s_c$ from $S_{I}$ such that the number of transitions in $\Theta_{b}^{1}$ is maximized
    \ENDIF
    \STATE Use the selected $s_c$ to construct the virtual transition sequence $\Theta_{v}$ as per Equation \ref{eqn:VTSfrombehandtargsequences}
    \STATE Use library $L_v$ to store the constructed sequence ($L_v=L_v\cup\{\Theta_{v}\}$)
    \ENDFOR
  \end{algorithmic}
  \label{alg:VTS}
\end{algorithm}

\subsection{Replaying the Transition Sequences}
\label{subsect3}
In order to make use of the transition sequences described, each of the
state-action-reward triads $\{s\quad a\quad r\}$ in the transition sequence
$L_{v}$ is replayed as if the agent had actually experienced them. 

Similarly, sequences in $L$ are also be replayed from time to time. Replaying sequences from $L$ and $L_{v}$ in this manner causes the
effects of large absolute TD errors originating from further up in the sequence to
propagate through the respective transitions, ultimately leading to more accurate
estimates of the value function.  The transitions are replayed as per the
standard $Q$-learning update equation shown below:
\begin{equation}
  \label{eqn:VTSupdate}
  Q(s_{j},a_{j})\leftarrow Q(s_{j},a_{j})+\alpha[R(s_{j},a_{j})+\gamma \max_{a'}Q(s_{j+1},a')-Q(s_{j},a_{j})].
\end{equation}
Where $s_{j}$ and $a_{j}$ refer to the state and action at transition $j$, and
$Q$ and $R$ represent the action-value function and reward
corresponding to the task. The variable $a'$ is a bound variable that represents any action
in the action set $\mathcal A$. The learning rate and discount parameters are
represented by $\alpha$ and $\gamma$ respectively.

The sequence $\Theta_{t_{s}}$ in Equation~\eqref{eqn:VTSfrombehandtargsequences} is
a subset of $\Theta_{t}$, which is in turn part of the library $L$ and thus
associated with a high absolute TD error.  When replaying $\Theta_{v}$, the
effects of the high absolute TD errors propagate from the values of
state/state-action pairs in $\Theta_{t_{s}}^{2}$ to those in $\Theta_{b}^{1}$.
Hence, in case of multiple points of intersection, we consider points that are
furthest down $\Theta_{b}$. In other words, the intersection point is chosen
to maximize the length of $\Theta_{b}^{1}$. In this manner, a larger number of
state-action values experience improvements brought about by replaying the
transition sequences.

\begin{algorithm}
  \caption{Replay of virtual transition sequences from library $L_{v}$}
  \begin{algorithmic}[1]
    \STATE \textbf{Inputs}: \STATEx $\alpha:$~learning rate \STATEx
    $\gamma:$~discount factor \STATEx
    $L_{v}=\{{\Theta_{v}}_{0}...{\Theta_{v}}_{i}...{\Theta_{v}}_{n_{v}}\}:$~A
    library of virtual transition sequences with ${n_{v}}$ sequences
   
    \FOR {$i=1:{n_{v}}$} 
    \STATE $n_{\text{\scriptsize sar}}=$number of $\{s\quad a\quad r\}$
    triads in ${\Theta_{v}}_{i}$ 
    \STATE $j=1$ 
    \WHILE {$j\leq n_{sar}$} \STATE
    $Q(s_{j},a_{j})\leftarrow
    Q(s_{j},a_{j})+\alpha[R(s_{j},a_{j})+\gamma
    \max_{a'}Q(s_{j+1},a')-Q(s_{j},a_{j})]$
    \STATE $j\leftarrow j+1$
    \ENDWHILE
    \ENDFOR
  \end{algorithmic}
  \label{alg:algorithm2}
\end{algorithm}

\section{Results and Discussion}
\label{results}
We demonstrate our approach on modified versions of two standard reinforcement
learning tasks. The first is a multi-task navigation/puddle-world problem (Figure
\ref{fig:description}), and the second is a multi-task mountain car problem (Figure
\ref{fig:description_mtcar}). In both these problems, behavior policies are
generated to solve a given task (which we refer to as the primary task)
relatively greedily, while the value function for another task of interest
(which we refer to as the secondary task) is simultaneously learned in an
off-policy manner. The secondary task is intentionally made more difficult by
making appropriate modifications to the environment.  Such adverse multi-task settings best demonstrate the effectiveness of our approach and emphasize its advantages
over
other experience replay approaches. We characterize the difficulty of the
secondary task with a difficulty ratio $\rho$, which is the fraction of the
executed behavior policies that experience a high reward with respect to the
secondary task.
A low value of $\rho$ indicates that achieving the secondary task under the
given behavior policy is difficult. In both tasks, the $Q-$ values are initialized with random values, and once the agent encounters the goal state of the primary task, the episode terminates.

\subsection{Navigation/Puddle-World Task}
\label{navigation}
\begin{figure}[ht]
  \centering
\includegraphics[width=0.65\linewidth]{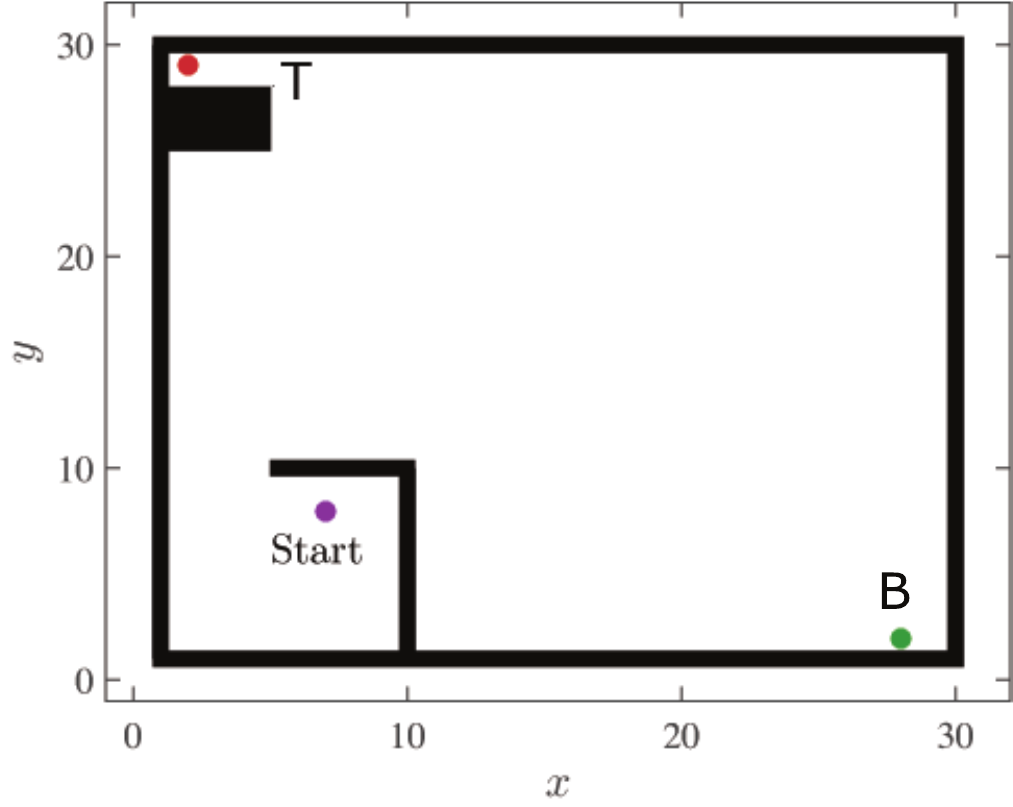}
  \caption{Navigation environment used to demonstrate the approach of
    replaying transition sequences}
  \label{fig:description}
\end{figure}

In the navigation environment, the simulated agent is assigned tasks of
navigating to certain locations in its environment. We consider two locations,
$B$ and $T$, which represent the primary and secondary task locations
respectively. The environment is set up such that the location corresponding
to high rewards with respect to the secondary task lies far away from that of
the primary task (see Figure~\ref{fig:description}). In addition to this, the accessibility to the secondary task
location is deliberately limited by surrounding it with obstacles on all but
one side. These modifications contribute towards a low value of $\rho$,
especially when the agent operates with a greedy behavior policy with respect
to the primary task.

The agent is assumed to be able to sense its location in the environment
accurately, and can detect when it `bumps' into an obstacle. It can move around in the environment at a maximum speed of 1
unit per time step by executing actions to take it forwards, backwards,
sideways and diagonally forwards or backwards to either side. In addition to
these actions, the agent can choose to hold its current position. However, the
transitions resulting from these actions are probabilistic in nature. The
intended movements occur only 80~\% of the time, and for the remaining 20~\%,
the $x$- and $y$-coordinates may deviate from their intended values by 1
unit. Also, the agent's location does not change if the chosen action forces
it to run into an obstacle.

The agent employs $Q$-learning with a relatively greedy policy
($\epsilon=0.1$) that attempts to maximize the expected sum of primary
rewards. The reward structure for both tasks is such that the agent receives a
high reward ($100$) for visiting the respective goal locations, and a high
penalty ($-100$) for bumping into an obstacle in the environment. In addition
to this, the agent is assigned a living penalty ($-10$) for each action that
fails to result in the goal state. In all simulations, the discount
factor $\gamma$ is set to be $0.9$, the learning rate $\alpha$ is set to a
value of $0.3$ and the parameter $\tau$ mentioned in Algorithm \ref{alg:algorithm1} is set to be $1$. Although various approaches exist to optimize the values of the $Q-$learning hyperparameters \citep{even2003learning,tutsoy2016analysis,garcia1998learning}, the values were chosen arbitrarily, such that satisfactory performances were obtained for both the navigation as well as the mountain-car environments.
\begin{figure}[ht]
  \centering
\includegraphics[width=0.7\linewidth]{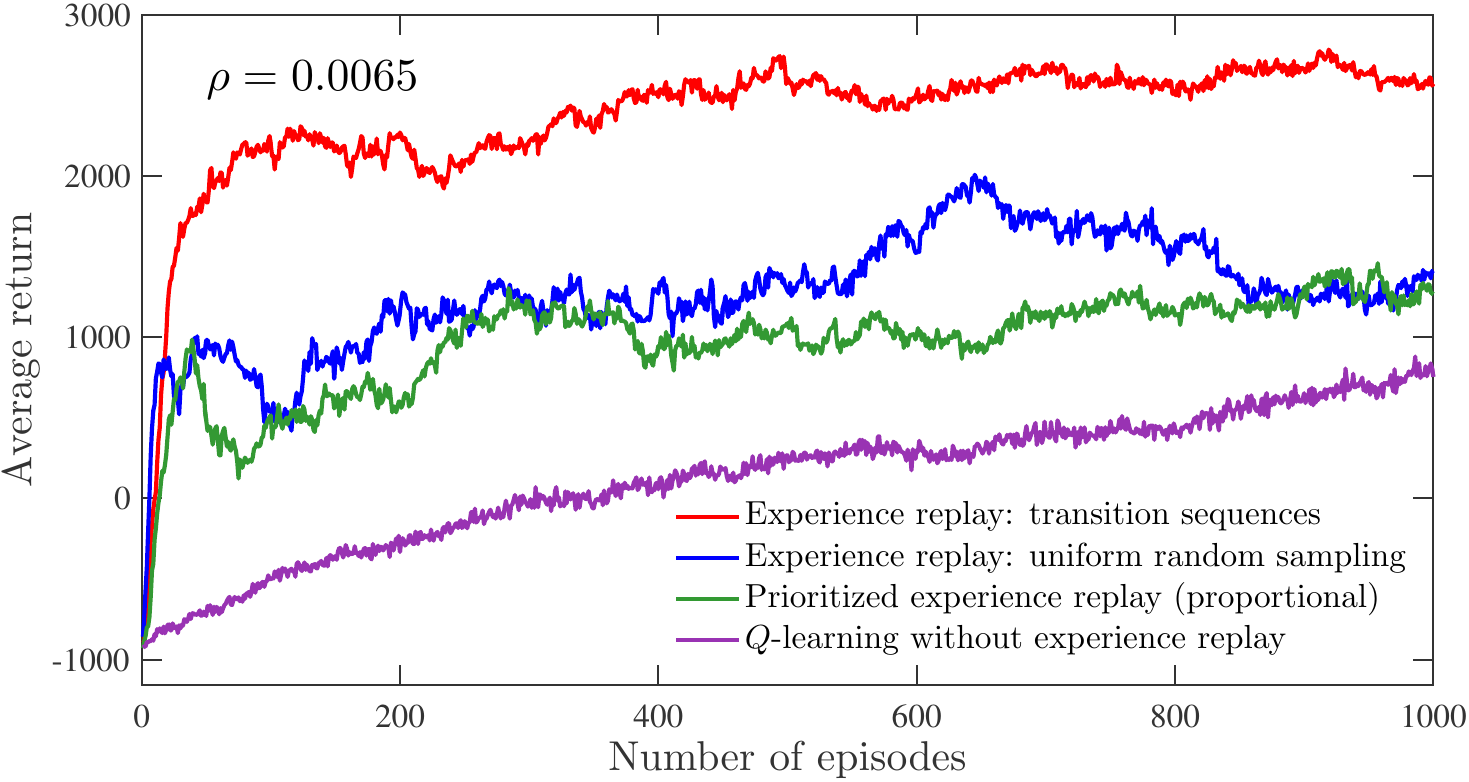}
  \caption{Comparison of the average secondary returns over $50$ runs using different experience replay approaches as well as $Q$-learning without experience replay in the navigation environment. The standard errors are all less than $300$. For the different experience replay approaches, the number of replay updates are controlled to be the same.}
  \label{fig:comparison}
\end{figure}

In the environment described, the agent executes actions to learn the primary
task. Simultaneously, the approach described in Section \ref{method} is
employed to learn the value functions associated with the secondary task. At
each episode of the learning process, the agent's performance with respect to
the secondary task is evaluated. In order to compute the average return for an episode, we allow the agent to execute $n_{ga}$($=100$) greedy actions from a randomly chosen starting point, and record the accumulated reward. The process is repeated for $n_{trials}$($=100$) trials, and the average return for the episode is reported as the average accumulated reward per trial. The average return corresponding to each episode in Figure \ref{fig:comparison} is computed in this way. The mean of these average returns over all the episodes is reported as $G_{e}$ in Table \ref{table1}.
That is, the average return corresponding to the $k^{th}$episode $g_{k}$ is given by: $$g_{k}=\frac{\sum\limits_{i=1}^{n_{trials}}\sum\limits_{j=1}^{n_{ga}}R_{ij}}{n_{trials}}$$ and $$G_{e}=\frac{\sum\limits_{k=1}^{N_{E}} g_{k}}{N_{E}}$$Where $R_{ij}$ is the reward obtained by the agent in a step corresponding to the greedy action $j$, in trial $i$, and $N_{E}$ is the maximum number of episodes.

Figure \ref{fig:comparison} shows the average return for the secondary task
plotted for $50$ runs of $1000$ learning episodes using different learning
approaches. The low average value of $\rho$ ($=0.0065$ as indicated in Figure
\ref{fig:comparison}) indicates the relatively high difficulty of the
secondary task under the behavior policy being executed. As observed in Figure
\ref{fig:comparison}, an agent that replays transition sequences manages to
accumulate high average returns at a much faster rate as compared to regular
$Q$-learning. The approach also performs better than other experience replay
approaches for the same number of replay updates. These replay approaches are applied independently of each other for the secondary task. In Figure \ref{fig:comparison}, the prioritization exponent for prioritized experience replay is set to $1$.

\begin{table}[!htb]
  \caption{Average secondary returns accumulated per episode ($G_{e}$) using
    different values of the memory parameters in the navigation
    environment}
  \label{table1}
  \begin{minipage}{.3\linewidth}
    \centering
    \caption*{(a)}
    \begin{tabular}{ll}
      $m_{b}$&$G_{e}$\\ \hline
      10 & 1559.7 \\ \hline
      100 & 2509.7  \\ \hline
      1000 & 2610.4   \\ \hline
    \end{tabular}
  \end{minipage}
  \begin{minipage}{.3\linewidth}
    \centering
    \caption*{(b)}
    \begin{tabular}{ll}
      $m_{t}$& $G_{e}$\\ \hline
      10 & 1072.5 \\ \hline
      100 & 1159.2 \\ \hline
      1000 & 2610.4 \\ \hline
    \end{tabular}
  \end{minipage}
  \begin{minipage}{.3\linewidth}
    \caption*{(c)}
    \centering
    \begin{tabular}{ll}
      $n_{v}$& $G_{e}$\\ \hline
      10 & 2236.6 \\ \hline
      50 & 2610.4 \\ \hline
      100 & 2679.5 \\ \hline
    \end{tabular}
  \end{minipage}%
  \begin{tablenotes}
    \small
  \item With regular $Q$-learning (without experience replay), $G_{e}=122.9$
  \end{tablenotes}
\end{table}

Table \ref{table1} shows the average return for the secondary task accumulated
per episode ($G_{e}$) during $50$ runs of the navigation task for different
values of memory parameters $m_{b}$, $m_{t}$ and $n_{v}$ used in our
approach. Each of the parameters are varied separately while keeping the other
parameters fixed to their default values. The default values used for $m_{b}$,
$m_{t}$ and $n_{v}$ are $1000$, $1000$ and $50$ respectively.
\subsubsection*{Application to the Primary Task}

\begin{figure}[ht]
  \centering
  \includegraphics[width=1\linewidth]{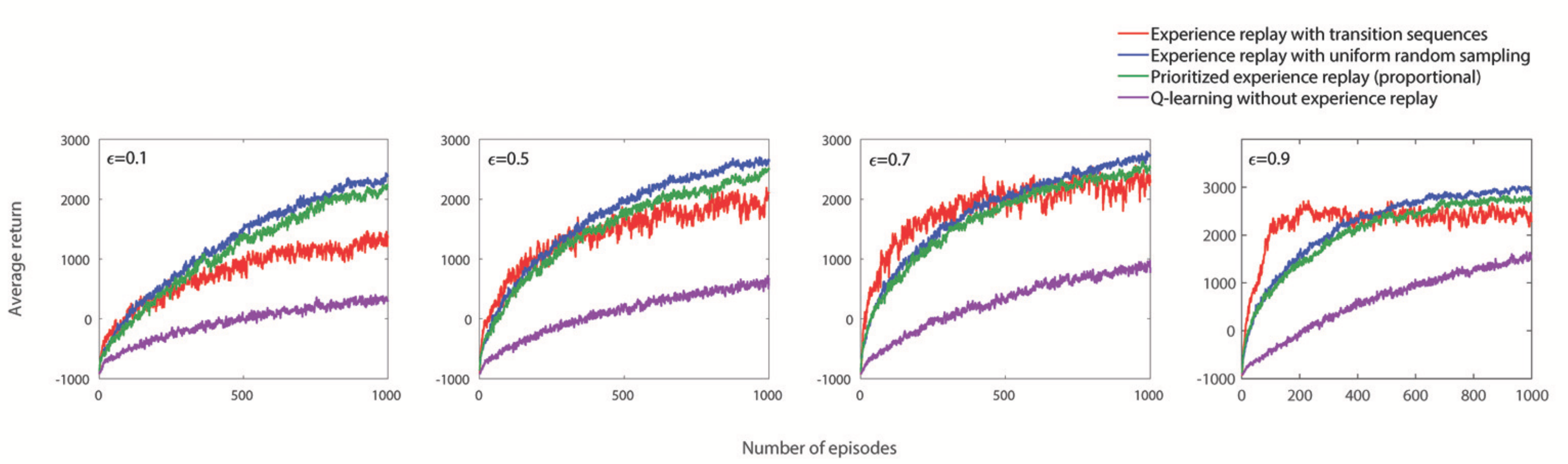}
  \caption{The performance of different experience replay approaches on the primary task in the navigation environment for different values of the exploration parameter $\epsilon$, averaged over $30$ runs. For these results, the memory parameters used are as follows: $m_{b}=1000$,
$m_{t}=1000$ and $n_{v}=50$.} \label{fig:prim_task}
\end{figure}

In the simulations described thus far, the performance of our approach was evaluated on a secondary task, while the agent executed actions relatively greedily with respect to a primary task. Such a setup was chosen in order to ensure a greater sparsity of high rewards for the secondary task. However, the proposed approach of replaying sequences of transitions can also be applied to the primary task in question. In particular, when a less greedy exploration strategy is employed (that is, when $\epsilon$ is high), such conditions of reward-sparsity can be recreated for the primary task. Figure \ref{fig:prim_task} shows the performance of different experience replay approaches when applied to the primary task, for different values of $\epsilon$. As expected, for more exploratory behavior policies, which correspond to lower probabilities of obtaining high rewards, the approach of replaying transition sequences is significantly beneficial, especially at the early stages of learning. However, as the episodes progress, the effects of drastically large absolute TD errors would have already penetrated into other regions of the state-action space, and the agent ceases to benefit as much from replaying transition sequences. Hence, other forms of replay such as experience replay with uniform random sampling, or prioritized experience replay were found to be more useful after the initial learning episodes. 

\subsection{Mountain Car Task}
\begin{figure}[ht]
  \centering
\includegraphics[width=0.7\linewidth]{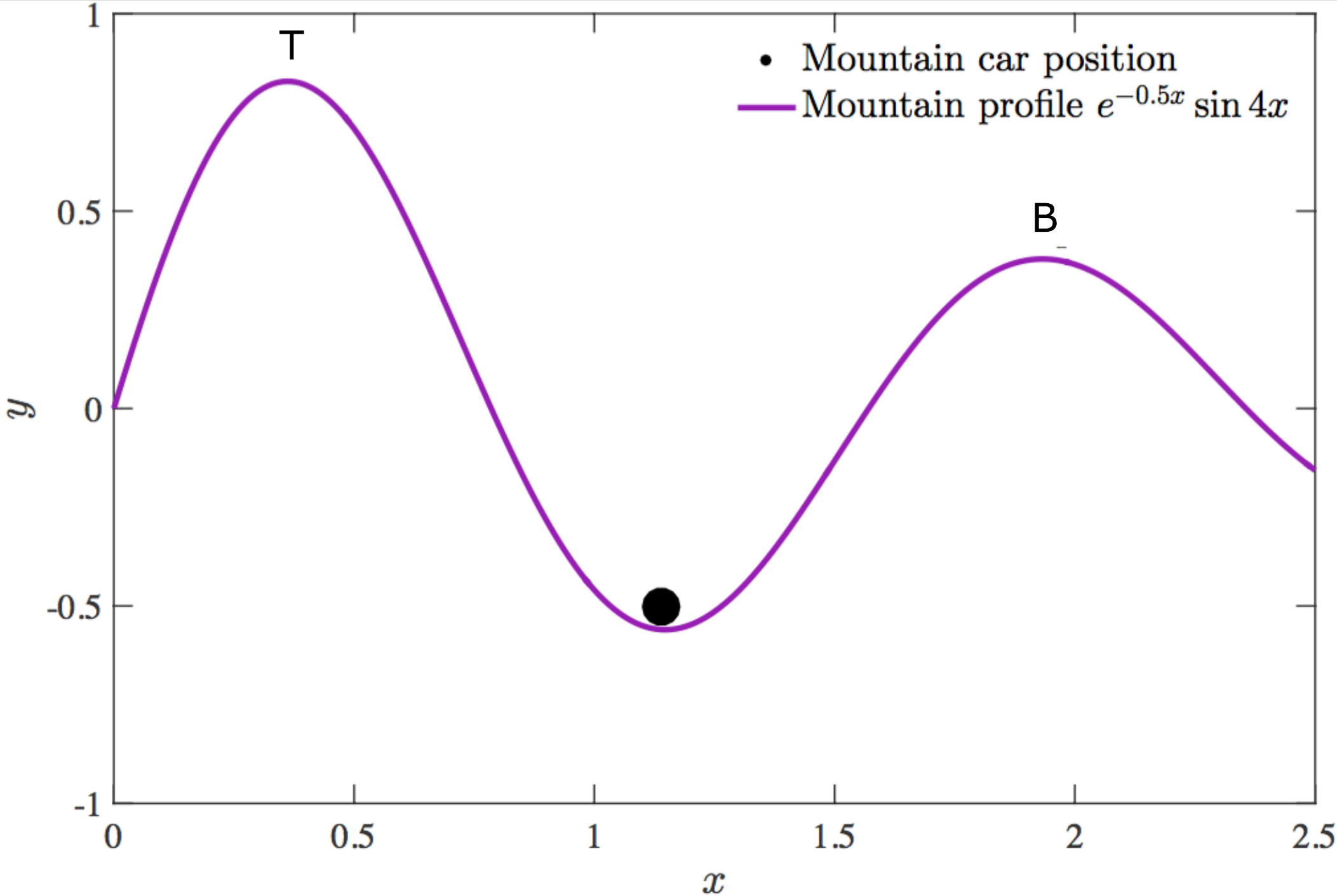}
  \caption{Mountain car environment used to demonstrate off-policy learning
    using virtual transition sequences}
  \label{fig:description_mtcar}
\end{figure}
In the mountain car task, the agent, an under-powered vehicle represented by
the circle in Figure \ref{fig:description_mtcar} is assigned a primary task of
getting out of the trough and visiting point $B$.  The act of visiting
point $T$ is treated as the secondary task. The agent is assigned a high
reward ($100$) for for fulfilling the respective objectives, and a living
penalty ($-1$) is assigned for all other situations.  At each time step, the
agent can choose from three possible actions: (1) accelerating in the positive
$x$ direction, (2) accelerating in the negative $x$ direction, and (3) applying no
control. The environment is discretized such that $120$ unique positions and $100$ unique velocity values are possible.

The mountain profile is described by the equation $y={e^{-0.5x}}\sin(4x)$ such that
point $T$ is higher than $B$. Also, the average slope leading to
$T$ is steeper than that leading to $B$. In addition to this, the
agent is set to be relatively greedy with respect to the primary task, with an
exploration parameter $\epsilon=0.1$. These factors make the secondary task
more difficult, resulting in a low value of $\rho$ ($=0.0354$) under the
policy executed.

\begin{figure}[ht]
  \centering
  \includegraphics[width=0.7\linewidth]{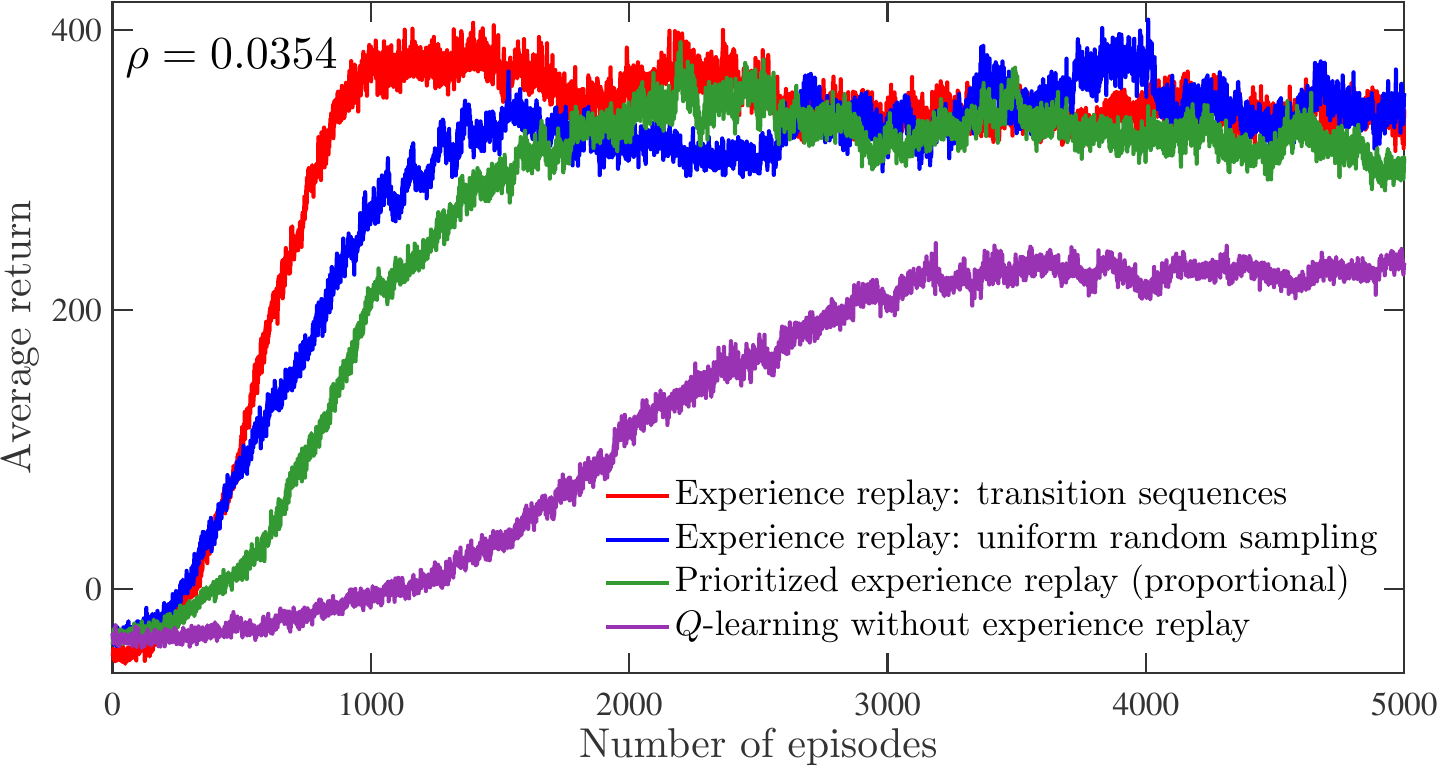}
  \caption{Comparison of the average secondary returns over $50$ runs using different
    experience replay approaches as well as $Q$-learning without experience
    replay in the mountain-car environment. The standard errors are all less than $85$. For the different experience
    replay approaches, the number of replay updates are controlled to be the
    same.} \label{fig:comparison_mtcar}
\end{figure}

Figure \ref{fig:comparison_mtcar} shows the average secondary task returns for
$50$ runs of $5000$ learning episodes. It is seen that especially during the
initial phase of learning, the agent accumulates rewards at a higher rate as
compared to other learning approaches. As in the navigation task, the number
of replay updates are restricted to be the same while comparing the different
experience replay approaches in Figure \ref{fig:comparison_mtcar}.
Analogous to Table \ref{table1}, Table \ref{table2} shows the average
secondary returns accumulated per episode ($G_{e}$) over $50$ runs in the
mountain-car environment, for different values of the memory parameters. The
default values for $m_{b}$, $m_{t}$ and $n_{v}$ are the same as those
mentioned in the navigation environment, that is, $1000$, $1000$ and $50$
respectively.

\begin{table}[H]
  \caption{Average secondary returns accumulated per episode ($G_{e}$) using different values of the memory parameters in the mountain car environment}
  \label{table2}
  \begin{minipage}{.3\linewidth}
    \centering
    \caption*{(a)}
    \begin{tabular}{ll}
      $m_{b}$&$G_{e}$\\ \hline
      10 & 221.0 \\ \hline
      100 & 225.1\\ \hline
      1000 & 229.9 \\ \hline
    \end{tabular}
  \end{minipage}
  \begin{minipage}{.3\linewidth}
    \centering
    \caption*{(b)}
    \begin{tabular}{ll}
      $m_{t}$& $G_{e}$\\ \hline
      10 & 129.9\\ \hline
      100 & 190.5\\ \hline
      1000 & 229.9 \\ \hline
    \end{tabular}
  \end{minipage}
  \begin{minipage}{.3\linewidth}
    \caption*{(c)}
    \centering
    \begin{tabular}{ll}
      $n_{v}$& $G_{e}$\\ \hline
      10 & 225.6 \\ \hline
      50 & 229.9\\ \hline
      100 & 228.4 \\ \hline
    \end{tabular}
  \end{minipage}%
  \begin{tablenotes}
    \small
  \item With regular $Q$-learning (without experience replay), $G_{e}=132.9$
  \end{tablenotes}
\end{table}

From Figures \ref{fig:comparison} and \ref{fig:comparison_mtcar}, the agent is
seen to be able to accumulate significantly higher average secondary returns
per episode when experiences are replayed.  Among the experience replay
approaches, the approach of replaying transition sequences is superior for the
same number of replay updates. This is especially true in the navigation
environment, where visits to regions associated with high secondary task
rewards are much rarer, as indicated by the low value of $\rho$. In the
mountain car problem, the visits are more frequent, and the differences
between the different experience replay approaches are less significant. The value of the prioritization exponent used here is the same as that used in the navigation task. The approach of replaying sequences of transitions also offers noticeable performance improvements when applied to the primary task (as seen in Figure \ref{fig:prim_task}), especially during the early stages of learning, and when highly exploratory behavior policies are used.
In both the navigation and mountain-car environments, the performances of the approaches that replay individual transitions---experience replay with uniform random sampling and
prioritized experience replay---are found to be nearly equivalent. We have not observed a significant advantage of using the prioritized approach, as reported in previous studies \citep{schaul2015prioritized,hessel2017rainbow} using deep RL. This perhaps indicates that improvements brought about by the prioritized approach are much more pronounced in deep RL applications.

The approach of replaying transition sequences seems to be particularly
sensitive to the memory parameter $m_{t}$, with higher average returns being
achieved for larger values of $m_{t}$. A possible explanation for this could
simply be that larger values of $m_{t}$ correspond to longer $\Theta_{t}$
sequences, which allow a larger number of replay updates to occur in more
regions of the state/state-action space.  The influence of the length of the
$\Theta_{b}$ sequence, specified by the parameter $m_{b}$ is also similar in
nature, but its impact on the performance is less emphatic.
This could be because longer $\Theta_{b}$ sequences allow a greater chance for
their state trajectories to intersect with those of $\Theta_{t}$, thus
improving the chances of virtual transition sequences being discovered, and of
the agent's value functions being updated using virtual experiences.  However,
the parameter $n_{v}$, associated with the size of the library $L_{v}$ does
not seem to have a noticeable influence on the performance of this
approach. This is probably due to the fact that the library $L$ (and
consequently $L_{v}$) is continuously updated with new, suitable transition
sequences (successful sequences associated with higher magnitudes of TD
errors) as and when they are observed. Hence, the storage of a number of
transition sequences in the libraries becomes largely redundant.
Although the method of constructing virtual transition sequences is more naturally applicable to the tabular case, it could also possibly be extended to approaches with linear and non-linear function
approximation. However, soft intersections between state trajectories would
have to be considered instead of absolute intersections. That is, while
comparing the state trajectories $S(x:y)$ and $S(x':y')$, the existence
of $s_{c}$ could be considered if it is close to elements in both $S(x:y)$
and $S(x':y')$ within some specified tolerance limit. Such modifications could allow the approach described here to be applied to deep RL. Transitions that belong to the sequences $\Theta_{v}$ and $\Theta_{t}$ could then be selectively replayed, thereby bringing about improvements in the sample efficiency. However, the experience replay approaches (implemented with the mentioned modifications) applied to the environments described in Section \ref{results} did not seem to bring about significant performance improvements when a neural network function approximator was used. The performance of the corresponding deep Q-network (DQN) was approximately the same even without any experience replay. This perhaps, reveals that the performance of the proposed approach needs to be evaluated on more complex problems such as the Atari domain \citep{mnih2015human}.
Reliably implementing virtual transition sequences to the function approximation case could be a future area of research. One of the
limitations of constructing virtual transition sequences is that in higher
dimensional spaces, intersections in the state trajectories become less
frequent, in general. However, other sequences in the library $L$ can still be
replayed. If appropriate sequences have not yet been discovered or
constructed, and are thus not available for replay, other experience replay
approaches that replay individual transitions can be used to accelerate
learning in the meanwhile. 

Perhaps another limitation of the approach described here is that constructing
the library $L$ requires some notion of a goal state associated with high
rewards. By tracking the statistical properties such as the mean and variance of the
rewards experienced by an agent in its environment in an online manner, the
notion of what qualifies as a high reward could be automated using suitable
thresholds \citep{KARIMPANAL201739}. In addition to this, other criteria such as the returns or average
absolute TD errors of a sequence could also be used to maintain the library.

\begin{figure}[ht]
  \centering
  \includegraphics[width=0.7\linewidth]{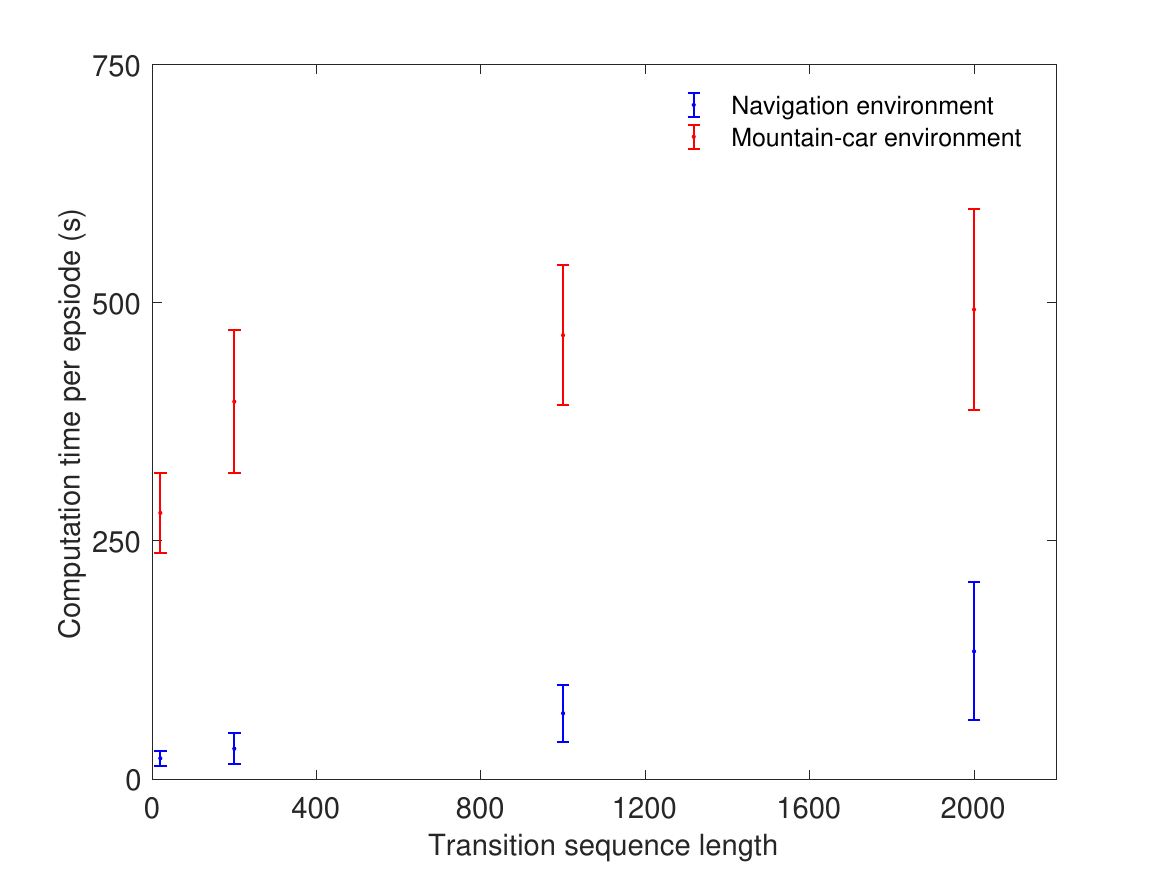}
  \caption{The variation of computational time per episode with sequence length for the two environments, computed over $30$ runs.} \label{fig:comput_time}
\end{figure}

It is worth adding that the memory parameters $m_{b}$, $m_{t}$ and $n_{v}$ have been set
arbitrarily in the examples described here. Selecting appropriate values for
these parameters as the agent interacts with its environment could be a topic
for further research. Figure \ref{fig:comput_time} shows the mean and standard deviations of the computation time per episode for different sequence lengths, over $30$ runs. The figure suggests that the computation time increases as longer transition sequences are used, and the trend can be approximated to be linear. These results could also be used to inform the choice of values for $m_{b}$ and $m_{t}$ for a given application. The values shown in Figure \ref{fig:comput_time} were obtained from running simulations on a computer with an Intel i7 processor running at 2.7 GHz, using 8GB of RAM, running a Windows 7 operating system. 

The approach of replaying transition sequences has direct applications in
multi-task RL, where agents are required to learn multiple tasks in
parallel. Certain tasks could be associated with the occurrence of relatively
rare events when the agent operates under specific behavior policies. The replay of
virtual transition sequences could further improve the learning in such
tasks. 
such as robotics, where exploration of the state/state-action space is
typically expensive in terms of time and energy. By reusing the agent-environment interactions in the manner described here, reasonable estimates of the value functions corresponding to multiple tasks can be maintained, thereby improving the efficiency of exploration.

\section{Conclusion}
In this work, we described an approach to replay sequences of transitions to
accelerate the learning of tasks in an off-policy setting. Suitable transition
sequences are selected and stored in a replay library based on the magnitudes
of the TD errors associated with them. Using these sequences, we showed that
it is possible to construct virtual experiences in the form of virtual
transition sequences, which could be replayed to improve an agent's learning,
especially in environments where desirable events occur rarely.
We demonstrated the benefits of this approach by applying it to versions of
standard reinforcement learning tasks such as the puddle-world and
mountain-car tasks, where the behavior policy was deliberately made drastically different from the target policy. In both tasks, a significant improvement in learning speed
was observed compared to regular $Q$-learning as well as other forms of
experience replay.  Further, the influence of the different memory parameters
used was described and evaluated empirically, and possible extensions to this
work were briefly discussed.
Characterized by controllable memory parameters and the potential to
significantly improve the efficiency of exploration at the expense of some
increase in computation, the approach of using replaying transition sequences
could be especially useful in fields such as robotics, where these factors are
of prime importance. The extension of this approach to the cases of linear and non-linear function approximation could find significant utility, and is currently being explored.


\section*{Acknowledgements}
This work is supported by the President's graduate fellowship (MOE, Singapore) and TL@SUTD under the Systems Technology for Autonomous Reconnaissance \& Surveillance (STARS-Autonomy \& Control) program. The authors thank Richard S. Sutton from the University of Alberta for his feedback and many helpful discussions during the development of this work. 

\bibliographystyle{elsarticle-num} 
\bibliography{test}

\end{document}